\documentclass[preprint,12pt]{elsarticle}

\usepackage{amssymb}
\usepackage{amsmath}
\usepackage{algorithmic}
\usepackage[]{algorithm2e}
\usepackage{graphicx}
\usepackage{adjustbox}
\usepackage{multirow}
\usepackage{subfigure}
\usepackage[table]{xcolor}
\usepackage[colorlinks=true, allcolors=blue]{hyperref}

\journal{Knowledge-Based Systems}

\begin{document}

\begin{frontmatter}



\title{Equation discovery framework EPDE: Towards a better equation discovery}


\author[NSS]{Mikhail Maslyaev}

\author[NSS]{Alexander Hvatov \corref{cor1}}
\ead{alex\_hvatov@itmo.ru}
\cortext[cor1]{Corresponding author}

\affiliation[NSS]{organization={NSS Lab},
            addressline={Kronverksky pr. 49}, 
            city={Saint-Petersburg},
            postcode={197101}, 
            country={Russia}}

\begin{abstract}
Equation discovery methods hold promise for extracting knowledge from physics-related data. However, existing approaches often require substantial prior information that significantly reduces the amount of knowledge extracted. In this paper, we enhance the EPDE algorithm -- an evolutionary optimization-based discovery framework. In contrast to methods like SINDy, which rely on pre-defined libraries of terms and linearities, our approach generates terms using fundamental building blocks such as elementary functions and individual differentials. Within evolutionary optimization, we may improve the computation of the fitness function as is done in gradient methods and enhance the optimization algorithm itself. By incorporating multi-objective optimization, we effectively explore the search space, yielding more robust equation extraction, even when dealing with complex experimental data. We validate our algorithm's noise resilience and overall performance by comparing its results with those from the state-of-the-art equation discovery framework SINDy.
\end{abstract}




\end{frontmatter}


\section{Introduction}

Originating from SINDy \cite{brunton2016discovering} and PDE-FIND \cite{rudy2017data}, equation discovery has evolved into a valuable machine learning tool for extracting knowledge from observational data. Extensions such as \cite{messenger2021weak} and \cite{fasel2022ensemble} have addressed the challenges posed by the experimental data. In SINDy, a predetermined term library is employed, incorporating a priori knowledge that the equation can be constructed using these terms. Often, the library is minimized to enhance its effectiveness. However, this approach raises concerns about the potential to extract unknown equations due to the limitations of the term library.

Using fixed functional structures in regression-based techniques, such as the form $u_t = F(x,u_x,u_{xx},...)$, tends to limit the discovery of entirely new equations, leading primarily to coefficient adjustments. This phenomenon extends to methods such as PDENet \cite{long2019pde}, PDE-READ \cite{stephany2022pde}, physics-informed neural networks (PINNs) \cite{raissi2019physics}, and DeepONet \cite{lu2021learning}. These approaches combine data and differential operators into a single loss function, which could introduce biases when addressing the challenge of discovering unknown equations. All these limitations are detailed in a survey \cite{camps2023discovering}.

We may propose evolutionary optimization as a more flexible yet slow alternative. Evolutionary optimization-based frameworks \cite{xu2020dlga, chen2022symbolic, maslyaev2021partial, atkinson2019data} offer an expanded exploration arena, facilitating dynamic term library generation. This broader search scope, encompassing an extended high-order linear term set and additional combinations, notably amplifies the optimization time. From the initial genetic algorithm introduced in \cite{atkinson2019data}, efforts have been made to accelerate computations and attain concise models. Techniques such as employing local regression with neural networks in the DLGA algorithm \cite{xu2020dlga}, sparse regression using candidate equation terms set in previous versions of EPDE \cite{maslyaev2021partial}, and representing each term as an independent computational graph within a forest in the SGA-PDE algorithm \cite{chen2022symbolic}, have been proposed to decrease the search space, thus accelerating the equation discovery process.

Effective equation discovery hinges not only on the equation representation but also on the optimization algorithm employed. For example, the superiority of multi-objective evolutionary algorithms over their single-objective counterparts is evident, even when these algorithms share a common objective \cite{maslyaev2023comparison}. Moreover, population-based algorithms present an advantage regarding dynamic search space alteration. To illustrate, using a term preference distribution can improve the stability and efficacy of the evolutionary algorithm search \cite{ivanchik2023directed}.

Comparative algorithm evaluation is complex due to several factors, including various noise quantification methodologies, disparate end-result assessment criteria, and variations in input data quantity. In particular, the intricacies are exemplified by the distinct treatments of noise levels across papers. For example, while some papers \cite{xu2023discovery, fasel2022ensemble} discuss noise at magnitudes of several hundred percent, others \cite{xu2021robust}, authored by the same researchers in some instances, report noise levels in the single digit percentages. This variance arises from nuanced noise application and computation approaches, often linked to reference points such as field mean or maximal value. Additionally, the evaluation intricacies are compounded by varying considerations such as the terms in the library, data and its derivatives, maximal equation order, and the maximum number of terms in the equation. Most importantly, code is not available to compare, except for a few examples mainly represented with SINDy and add-ons \cite{Kaptanoglu2022}.

This paper introduces several key advancements within the emerging EPDE framework \cite{maslyaev2021partial}, delineating key directions that culminate in a substantial progression. These innovations collectively enable a notable improvement in noise robustness compared to their counterparts. The outlined advances are as follows:

\begin{itemize}
    \item  \textbf{Transition to multi-objective optimization} We change from the conventional single-objective evolutionary optimization paradigm to the multi-objective counterpart. This transition is advantageous, even when the ultimate goal is a singular objective (such as an error metric).

    \item \textbf{Enhanced handling of equation systems} Within the framework of multi-objective optimization, we refine the handling of equation systems encompassing Ordinary Differential Equations (ODEs) and Partial Differential Equations (PDEs). This refined approach contributes to the overall efficacy of the framework.

    \item \textbf{Improved treatment of differentials} We present an improved methodology for dealing with differentials absent from the input data. This involves a combination of neural network interpolation and subsequent numerical or automatic differentiation, bolstering the framework's adaptability.

    \item \textbf{Optimal management of input hyperparameters} We introduce a streamlined approach to managing input hyperparameters, thus facilitating the seamless exploration of unknown ODE and PDE discovery. This versatility is achieved through judicious parameter selection.
\end{itemize}

We compare our approach with the pySINDy framework \cite{Kaptanoglu2022}, a standard in the field, using the data provided by the pySINDy authors as examples and numerical equation solutions. Our findings indicate that stochastic algorithms like EPDE, which facilitate dynamic term library generation, notably surpass deterministic regression methods such as SINDy for discovering unknown equations. This capability helps to distinguish noise terms from significant ones, contributing to superior noise robustness. However, this enhancement is traded off with longer optimization times. To ensure reproducibility, we provide code for both SINDy and EPDE experiments.

The paper is organized as follows. Section~\ref{sec:alg_descr} is dedicated to the description of our approach and combines Subsection~\ref{subsec:representation}, which contains ideas about how we handle the encoding of differential equations; Subsection~\ref{subsec:criteria} is aimed at examining the question of selecting the optimization criteria, and Subsection~\ref{subsec:criteria} describes the multi-objective evolutionary optimization approach. Section~\ref{sec:validation} provides the validation of the method compared to the state-of-the-art competitor. In Subsections~\ref{subsec:PDE_validation}, \ref{subsec:ODE_validation} and \ref{subsec:system_validation}, we examine the performance of the algorithms on noisy data for partial and ordinary differential equations and systems of differential equations. We conclude the research in Section~\ref{sec:concl}.

\section{Overview of the proposed method}
\label{sec:alg_descr}

\begin{figure}[ht!]
\centering
\includegraphics[width=0.99\textwidth]{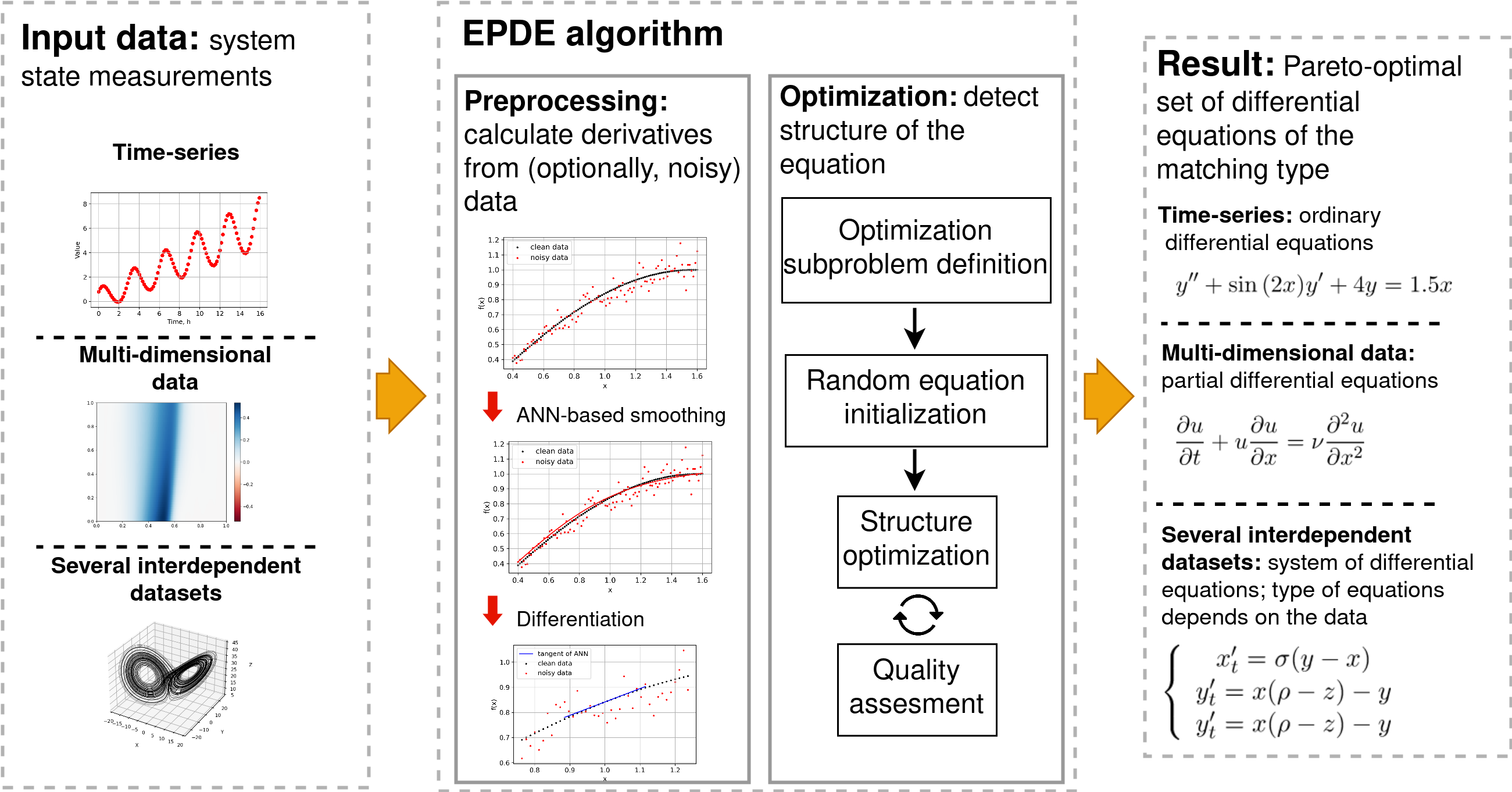}
\caption{The workflow of the proposed approach for the evolutionary-based data-driven algorithm of differential equations discovery. 
}
\label{fig:sketch}
\end{figure}

Generally, in the approach based on the evolutionary algorithm, discovering differential equations can be reinstated as the construction of expression, connecting various elementary function derivatives, which is optimal from the standpoint of selected criteria on the input data. The main idea of our approach is provided in scheme Fig.~\ref{fig:sketch}

It is assumed that the observational data $\mathbf{u}(\mathbf{x})=\{u_1(\mathbf{x}),...,u_k(\mathbf{x})\}$ have the form of a grid function on a discrete grid $\mathbf{x}={x_1,...,x_{dim}}$, where $dim$ is the dimensionality of the problem, $x_i$ are parameters (spatial and temporal coordinates, particle sizes) and $k$ is the number of observed values. For the differential equation discovery problem, it is also assumed that the function represented by the observational data $\mathbf{u}(\mathbf{x})=\{u_1(\mathbf{x}),...,u_k(\mathbf{x})\}$ is differentiable up to the maximal order of the differential equation that the algorithm can obtain.

To define the equation discovery algorithm, one should consider three aspects of the problem, which are: 

\begin{itemize}
\item approach to the definition of elementary functions and equation encoding, 
\item setting of criteria, according to which the optimization will be handled,
\item the algorithm that will perform the optimization. 
\end{itemize}

All three points are discussed in the following subsections from the EPDE framework point of view.

\subsection{Representation of differential equations}
\label{subsec:representation}

In the discovery process, the elementary functions $t_j = t_j(p_{j 1}, p_{j 2}, \; ...)$ are used as factors in the equations. For convenience, they are grouped into parametrized classes (families). The basic class (family) of elementary functions is the input variable derivatives in the form of $F_{deriv} = \{\frac{\partial^{n} u}{\partial x_1 ^{n_1}\; ... \; \partial x_{dim}^{n_k}}\}$, with the $n_1 + \; ... \; + n_k = n \in {1, 2, \; ... \;, N_d}$. With $N_d$, we denote the hyperparameter of the maximal order of the equation. The equation may be of lesser order $N \le N_d$. All selected families of elementary functions are combined into a pool $F = F_{deriv} \cup F_1 \cup \; ...$, from which factors are chosen during the randomized equation generation process and in evolutionary operators.

Equation factors are combined into terms by product capped by the number of terms hyperparameter $N_{terms}$. The equation may contain fewer terms $N \le N_{terms}$. Token families may be marked as ``autonomous'' to form non-homogeneity. They are not necessarily chosen as part of other terms. Such autonomous tokes and the family of derivatives form the set $F_{indep}$. The resulting terms are linearly combined to form the differential equation as in Eq.~\eqref{eq:candidate_pde}.

\begin{equation}
\label{eq:candidate_pde}
\begin{array}{cc}
    L'u = \sum^{N_{terms}}_{i = 0} a_i(\mathbf{x}) c_i(\mathbf{x})  = 0 \; ,\\ 
    a_i(\mathbf{x}) = b_i*a'_i(\mathbf{x}) = b_i*\prod_{j} t'_{ij}(\mathbf{x}), \; b_i \in \mathbb{R}, \; t'_{ij} \in F \setminus F_{indep} \\
    c_i(\mathbf{x}) = \prod_j t_{ij}(\mathbf{x}), \; t_{ij} \in F_{indep}
\end{array}
\end{equation}

While specifying token families can be automated, the default approach involves semi-manual selection. Baseline sets must include the derivatives $F_{deriv}$. In case of processing data in the dynamical environment, such as heat transfer in the medium affected by convection, the set can contain the family of vector velocity components $F_{velocity} = \{v_1(\mathbf{x}), v_2(\mathbf{x}), \; ...\}$. In an ideal case, the velocity is measured and contained in the input data, although in other cases, it can be represented as the parametric function, subject to optimization. Other typical token families include inverse functions of grid arguments $F_{inv} = \{\frac{1}{x_i}, \; i = 1, \; ... \; , dim$\}, polynomial functions of the grid arguments up to a defined order $N$ $F_{grid\_poly} = \{\sum_{j=0}^{n} p_{j} x_{i}^{j}, n = 1, \; ... \;, N; \; i = 1, \; ... \; , dim$ \}, or polynomials of the dependent variable $F_{var\_poly} = \{\sum_{j=0}^{n} p_{j} u^{j}, n = 1, \; ... \;, N\}$.

For the evolutionary algorithm, a system of differential equations can be represented in a list-like chromosome, where the elements represent single equations of the system and introduced parameters of equation construction, marked as available for optimization. Each equation is encoded through the tree graph structure: the root is the summation operator, as in Eq.\eqref{eq:candidate_pde}. The lower node level represents multiplication operations between the factors inside the term, whereas the leaves comprise tokens, excluding real-valued coefficients.

When operating with differential equation systems $S(\mathbf{u}) = 0$, each of the individual equations $L_i(\mathbf{u})$ of the system is forced to describe (partially) a specific variable $u_i$, that is, to contain its derivative. However, an equation of the system is not restricted from containing derivatives of the other variables or tokens.

\begin{equation}
    S(\mathbf{u})=\left\{\begin{array}{cc}
    
         L_1(\mathbf{u})=0  \\
         ... \\
         L_k(\mathbf{u})=0
    \end{array}
    \right.
\label{eq:statement_system}
\end{equation}

The parameters of the differential equation algorithm include the number of terms and the number of factors in the terms. These parameters determine the initial term definition during equation generation.

\subsection{Optimization criteria}
\label{subsec:criteria}

The desire to control the complexity of the obtained models and minimize the error manifests itself in making preferences for parsimonious equations. In practice, the parsimony of data-driven differential equations is connected to the number of terms in the equation structure, the order of derivatives inside of them, and its process representation ability. The criteria for equation discovery used in the algorithm include the complexity metric Eq.~\eqref{eq:norm_complexity}, where by $n$ we define the order of the derivative. Corresponding tokens are given a base complexity of 0.5 to avoid equation overfitting (for example, creating overly complex non-homogeneity) with elementary functions not represented by derivatives.

\begin{equation}
\begin{array}{cc}
      C (L'u) = \sum_{j}\sum_{i} \text{compl} (t_{ij}); \\
      \text{compl}(t_{ij}) = \begin{cases} n &, \; if \; t_{ij} = \frac{\partial^{n} u}{\partial^{n_1} x_1 \; ... \; \partial^{n_k} x_{dim}}, n \geq 1 \\ 0.5 &, \; \text{otherwise} \end{cases}
\end{array}
\label{eq:norm_complexity}
\end{equation}

We propose using error metrics to evaluate the quality of a candidate equation. The first one includes an estimation of the discrepancy of the differential equation operator. Under ideal conditions, the equation represents an identity of the expression containing unknown modeled functions and their derivatives to zero. While trivial combinations, utilizing zero coefficients, of any predictors can form a correct identity in a homogeneous equation case, they present no interest for the model creation. To avoid convergence to a zero coefficient vector, it is feasible to designate a random candidate term with index $target\_idx$  (left-hand side) to have a fictive coefficient of $b_{target\_idx} = -1$, so the problem is restated as the detection of the optimal real-valued coefficients for the other terms (right-hand side) with respect to this condition, as in Eq.~\eqref{eq:coeff_search}.

\begin{multline}
\mathbf{b}_{opt} = \text{arg} \min_{\mathbf{b}} ||L'u|| =\\
= \text{arg} \min_{\mathbf{b}} \Big|\Big| \sum^{N_{terms}}_{i = 0, \; i \neq target\_idx} b_i a'_i(\mathbf{x}) c_i(\mathbf{x}) + b_{bias} - c_{target\_idx}(\mathbf{x}) \Big|\Big|
\label{eq:coeff_search}
\end{multline}

With the notion that the discrepancies in the center of the domain are more informative due to errors in derivatives estimation near the domain boundaries and the possibility of other interfering processes, we can introduce weight function $g(\mathbf{x})$ and vector of adjusted coefficients $\widetilde{\mathbf{b}}_{opt}$, where $b_{bias}$ is the constant part of the equation non-homogeneity. The optimization problem is restated in Eq.~\eqref{eq:coeff_search_domain}:

\begin{equation}
\widetilde{\mathbf{b}}_{opt} = \text{arg} \min_{\mathbf{b}} || (\sum^{n\_terms}_{i = 0, \; i \neq target\_idx} b_i a'_i(\mathbf{x}) c_i(\mathbf{x}) + b_{bias} - c_{target\_idx}(\mathbf{x})) \cdot g(\mathbf{x}) ||
\label{eq:coeff_search_domain}
\end{equation}

In general, during optimization for problems in Eq.~\eqref{eq:coeff_search} and Eq.~\eqref{eq:coeff_search_domain} an arbitrary norm $|| \cdot ||$ can be used. Since the input data for the algorithm consists of measurement arrays representing variable values on the grid, it is natural to utilize the L2-norm of the vectors of discrepancies evaluated in each node. The values of the terms are combined into vectors: for the $i$-th one, $F_i = (a'_i(\mathbf{x}_{l}) c_i(\mathbf{x}_{l}), \; ... \; , a'_i(\mathbf{x}_{l}) c_i(\mathbf{x}_{l}))^{T}$ of its values on the grid nodes, indexed by $l \in \Lambda$, the grid indexing set. All terms except the designated one are combined in Eq.~\eqref{eq:feature_matrix} into the feature matrix $\mathbf{F}$.

\begin{equation}
\mathbf{F} = 
\begin{pmatrix}
| &  & | & | &  & | \\
F_{1} & ... & F_{target\_idx - 1} & F_{target\_idx + 1} & ... & F_{n\_terms} \\
| &  & | & | &  & | 
\end{pmatrix}
\label{eq:feature_matrix}
\end{equation}

The problem statement matches the weighted linear regression problem of Eq.~\eqref{eq:linear_regression}: the vector of designated term values $F_{target\_idx}$ can be selected as the target for the approximation, while the matrix $\mathbf{F}$ of the remaining terms represents the predictors. The weights form the diagonal matrix $G$. There, in element $G_{kk}$, the value of $g(\mathbf{x}_k)$ - weight function in the grid node, matching the $k$-th index of the set $\Lambda$, is located. 

\begin{equation}
\label{eq:linear_regression}
    \widetilde{\mathbf{b}}_{opt} = \text{arg} \min_{\mathbf{b}} \Vert{G^{\frac{1}{2}}(\mathbf{F}}\mathbf{b} - F_{target\_idx})\Vert_{2}^{2}
\end{equation}

The value of the L2-norm of the discrepancy between $F_{target\_idx}$ and the sum of the other terms $F_{i}$ can be used as the criterion for optimizing the structure of the equation, and the corresponding fitness function is represented by Eq.~\eqref{eq:fitness_val_old}. In practice, the constructed identity does not hold even with the correct structure and the correct coefficients due to numerical errors of the derivative calculation and inaccuracies of the coefficient estimation with linear regression.    

\begin{equation}
    Q_{diff\_op}(L'u) = (|| \sum_{i \neq i\_rhs} (a^{*}_i(t, \mathbf{x}) b_i c_i) - a^{*}_{i\_rhs}(t, \mathbf{x}) c_{i\_rhs} ||_2)
    \label{eq:fitness_val_old}
\end{equation}

An alternative approach to estimating the quality of equations involves comparing the equation/system of equations solution $\widetilde{\mathbf{u}}$ to the input data $\mathbf{u}$:

\begin{equation}
    Q_{sol}(L'u) = (|| \widetilde{u}(\mathbf{x}) - u(\mathbf{x}) ||_2)
    \label{eq:fitness_val_solver}
\end{equation}

An interface is implemented with the PINN-based approach \cite{hvatov2023automated} to solve equations appearing in the search process. Classic approaches, such as finite differences or finite elements, can only be applied if preliminary assumptions about the equation form exist. The proposed equation must be solved according to the automatically specified initial/boundary problem, where the solution domain corresponds to the observation data grid, and the type of constructed equations defines the conditions. As in the conventional analytic approach to solving differential equations, in the PINN-based method, the result is a form of a function of the independent variables (coordinates).

Although using the automated search to detect the solution function in the free form is not viable, the parametric solution function $\widetilde{u}(\mathbf{x}, \Theta)$ can be relatively easily optimized. In the case of a fully connected deep neural network representation of the solution, this optimization will be maintained as the search for the parameters $\Theta$ in neurons with backpropagation with the loss function Eq.~\eqref{eq:optimization_problem}. A generalized loss function contains a term describing the fitting of the function to the differential operator and a term representing the fitting boundary operator, weighted by $\lambda$ constant.

\begin{equation}
\label{eq:optimization_problem}
    (|| L \widetilde{u}(\mathbf{x}) - f ||_i + \lambda ||G \widetilde{u}(t, \mathbf{x}) - g||) \longrightarrow \min_{\widetilde{u}}
\end{equation}

The automatized approach involves initialization with discretized differential operators $\widetilde{L}$ and $G$, obtained with finite differences. Partial derivatives are replaced by their discrete equivalents. For example, the first-order partial derivatives with respect to the $i$-th independent variable are approximated as in Eq.~\eqref{eq:fin_diff_c}, where $\delta_{i}$ is the vector of coordinate increment $\mathbf{\delta^{i}}: \delta^{i}_{j} = 0 \; \text{if} \; i \neq j$, and the $i$-th component $\delta^{i}_{i}$ step of the corresponding grid.

\begin{equation}
\label{eq:fin_diff_c}
    \frac{\partial \widetilde{u}(\mathbf{x})}{\partial t} = \frac{\widetilde{u}(\mathbf{x} + \mathbf{\delta^{i}}) - \widetilde{u}(\mathbf{x} - \delta^{i})}{2 * \delta^{i}_{i}}
\end{equation}

Similarly, the discretized initial or boundary conditions operators are introduced. Evaluation of solution-based fitness functions requires constructing such operators from the data. At the same time, the numerical specification of non-Dirichlet boundary conditions is complicated by the necessity to compute derivatives of the observation data field. The derivative values, numerically obtained by analytical differentiation of polynomials, approximating the data on a subdomain or by automatic differentiation of neural network-based representation, are utilized to define the demanded operators.

Although the solution-based approach has high computational costs, it is less prone to guiding the algorithm's convergence to the incorrect form of noisy data. Therefore, the framework should decide the preferential criteria: if the data is expected to have high noise magnitudes, the solver-based approach, while if the measurements are relatively accurate, the discrepancy-based method can be used.

\subsection{Evolutionary optimization}
\label{subsec:optimization}

For single equation discovery with objective criteria $Q(L'u)$ and $C(L'u)$, introduced in the previous section, the problem of multi-objective optimization is stated as the search for the correct terms of the equation $\{a'_i, c_i\}, i = 1, \; ... \; N_{terms}$, paired with real-valued coefficients \& bias coefficient $\{b_i\}, i = 1, \; ... \; N_{terms+1}$ in respect to the metrics $Q(L'u)$ and $C(L'u)$. Previous research has proved that the multi-objective approach, where optimization is held simultaneously for both metrics, has faster and more reliable convergence to the true governing equation than the single-objective approach. In case of algorithm operation on the data, describing processes that involve multiple variables, complexity and quality metrics are introduced for each equation individually, as $Q_{i}(S(\mathbf{u})) = Q(L_{i}'u)$, $C_{i}(S(\mathbf{u})) = C(L_{i}'u)$. The general vector of objectives can be constructed as $\mathbf{F}(S(\mathbf{u})) = (Q_{1}(S(\mathbf{u})), C_{1}(S(\mathbf{u})), \; ... ,$ $\; Q_{n\_var}(S(\mathbf{u})), C_{n\_var}(S(\mathbf{u})))$.

For two candidate systems, the dominance condition can be introduced: we say that the system $S_1(\mathbf{u})$ dominates the system $S_2(\mathbf{u})$ (and is denoted as $S_1(\mathbf{u}) \preceq S_2(\mathbf{u})$), if for every equation $i$ in the system $Q_{i}(S_{1}(\mathbf{u})) \leq Q_{i}(S_{2}(\mathbf{u}))$, and $C_{i}(S_{1}(\mathbf{u})) \leq C_{i}(S_{2}(\mathbf{u}))$, there exists equation $j$, for that $Q_{j}(S_{1}(\mathbf{u})) < Q_{j}(S_{2}(\mathbf{u}))$ and/or $C_{j}(S_{1}(\mathbf{u})) < C_{j}(S_{2}(\mathbf{u}))$. A set of candidate systems of the equation is called non-dominated if, for any two solutions on the set, we can not say that any one dominates. The implemented discrete optimization aims to detect the Pareto-optimal non-domination set containing solutions over which no other solutions dominate.

To divide the population into non-domination levels, we employ the tool proposed by \cite{deb2002fast}.

The multi-objective evolutionary optimization approach, based on the dominance and decomposition proposed in \cite{li2014evolutionary}, is employed to search the set of Pareto-optimal equations/systems of equations. The decomposition method involved the use of $g^{pbi}(S(\mathbf{u}) | \mathbf{w}_i)$, the penalty-based intersection (PBI) value, where $\mathbf{w}_i$ represents a weight vector, representing a particular sub-problem.
The penalty-based intersection comprises two components: $d_1$, representing the measure of solution convergence towards the optimal systems, and $d_2$ representing the diversity of the population, promoting solutions co-directional to the weight vector.

\begin{equation}
\label{eq:decomposition}
\begin{split}
    g^{pbi}&(S(\mathbf{u}) | \mathbf{w}_i ) = d_1 + \theta d_2 \\
    d_1 &= \frac{\Vert (\mathbf{F}(S(\mathbf{u})))^{T} \; \mathbf{w}_i \Vert}{\Vert \mathbf{w}_i \Vert} \\
    d_2 &= \Vert \mathbf{F}(S(\mathbf{u})) - d_1 \frac{\mathbf{w}_i}{\Vert \mathbf{w}_i \Vert} \Vert
\end{split}
\end{equation}

While the equation search problem poses a number of limitations on the structures, none are tackled directly as the constraints during the evolutionary optimization. The operation of the algorithm can be decomposed into the following steps:

\textit{Estimation of the best achievable criteria values.} Both introduced objective functions should be minimized: for the quality metric value $Q(L'u)=0$, it matches an equation that completely governs the process. The best case of the complexity metric $C(L'u) = 0$ corresponds to the equation $\frac{\partial u}{\partial t} = const, \; const \in \mathbb{R}$, which, of course, does not describe most of the physical processes. However, for the complexity metric, we still use the physical sense that the equation should be as simple as possible. As a result of such an ideal solution definition, the algorithm aims to detect the most simple yet descriptive equation and should not overtrain, that is, to create a complex model for an elementary process.

Partition of the criteria value space with the set of weight vectors: $W = \{\mathbf{w}_1, \mathbf{w}_2, \; ... \;, \mathbf{w}_{n\_pop}\}$, where each weight vector is introduced to represent an optimization sub-problem (sub-region $\Phi_1, \Phi_2, \; ... \; , \Phi_{n\_pop}$), for which a solution is tried to be discovered. The decomposition does not rigidly separate the population but promotes an even distribution of the solutions in the space. Therefore, the cardinality of the set should match the defined number of candidate solutions in the population.

Each weight vector $\mathbf{w}_1$ is sampled from the unit simplex so that the spacing is equal between the vectors. The number of candidates and weight vectors must be sufficiently high to properly explore the criteria value space. In the original article \cite{DasDennis1998}, the authors recommend introducing $n\_pop = \binom{H + m - 1}{m - 1}$, where $m$ is the number of criteria and $H$ is the number of divisions of the criteria axis, with the proposition of $H \geq m$. However, the computational cost of processing a single individual is too high for realistic execution of the search process with a high number of individuals, especially in problems of systems of differential equations. Therefore, fewer divisions must be made, reducing the density of the criteria space coverage and decreasing diversity.

\textit{Random creation of the initial population} $P = \{ S_1(\mathbf{u}), S_2(\mathbf{u}), \; ... \; , S_{n\_pop}(\mathbf{u}) \}$ of candidate equations (systems). In addition to creating an equation representation graph, the sparsity constant value $\lambda$ is associated with the structure of the equation to control the promotion of sparsity. Each constructed equation is randomly associated with the sub-region, defined by a weight vector to represent the solution of an optimization sub-problem. Most interactions are held with the neighbouring sub-regions, estimated using Euclidean distances between the weight vectors.

\textit{Reproduction cycle.} The evolutionary approach to creating new candidate solutions combines two operators: cross-over and mutation. For each of the weights paired with the corresponding candidate solution, the reproduction is held on the execution epoch of the evolutionary algorithm. The mating selection operator holds the selection of equations/systems to be used as the parent individuals. With a defined probability $\delta$ $k$-parent solutions will be chosen from the sectors located in the sector and its neighbours (if there are any solutions), while in the opposite case, the parents are selected randomly from the entire population. The first selection pattern is introduced to exploit existing solutions that describe the domain. On the contrary, the second is used to explore new solutions. The selected individuals are split into pairs, and the cross-over is held among them.

\begin{figure}[ht!]
\centering
\includegraphics[width=0.85\textwidth]{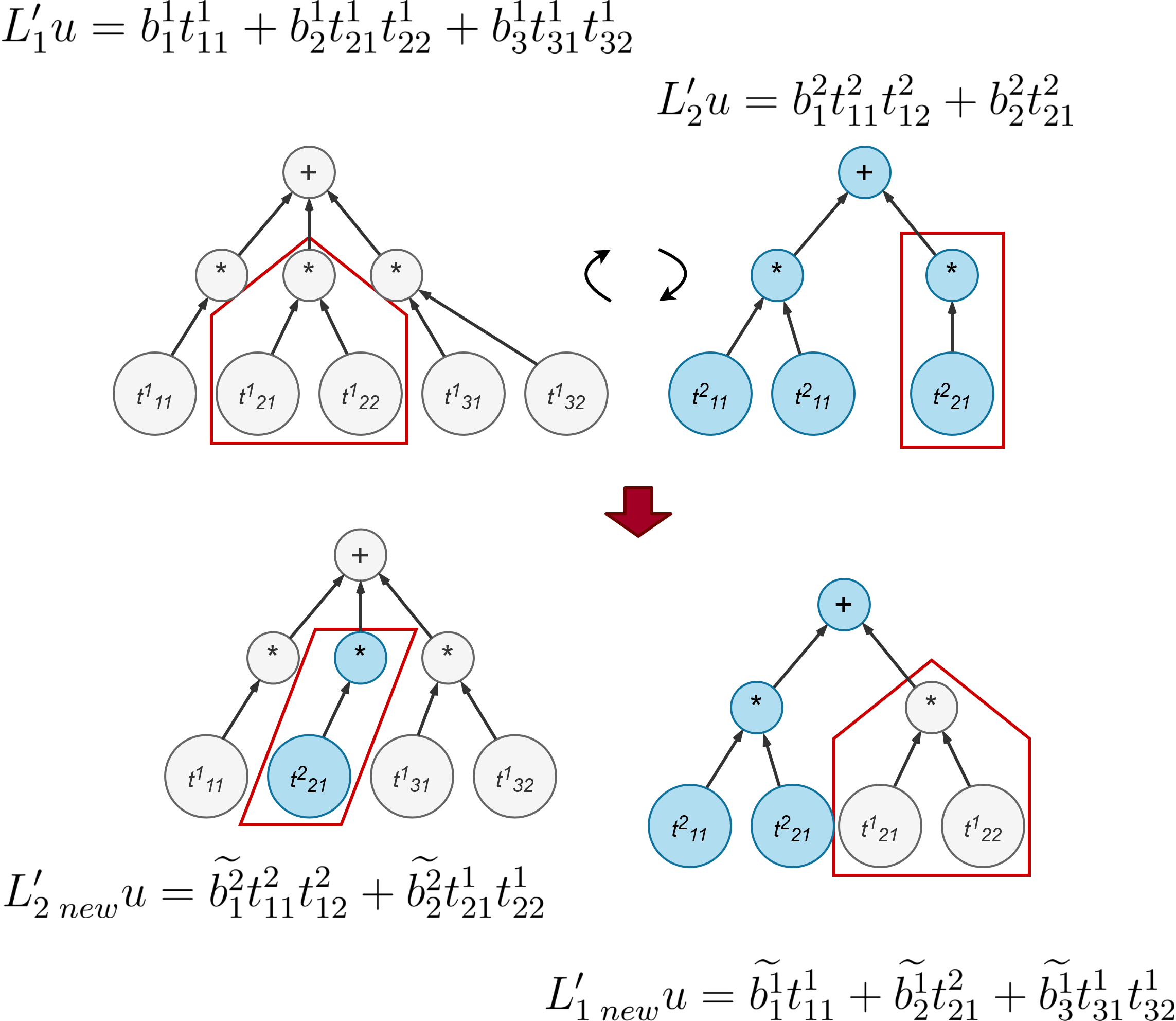}
\caption{Scheme of the cross-over operator.}
\label{fig:cross-over}
\end{figure}

The cross-over operator is based on the interactions between the structures representing the equations describing the same variable. Each pair of parents produces two offspring. Two types of operators are introduced since the encoding contains the symbolic graphs of the equations and the values of the sparsity constants. For the sparsity constant, the new values are selected in the interval between the parent values $\lambda'_{1} = \alpha \lambda_{1} + (1 - \alpha) \lambda_{2}$ and $\lambda'_{2} = (1 - \alpha) \lambda_{1} + \alpha \lambda_{2}$.

For the chromosomes representing the individual equations of the system, we can introduce a graph exchange operator, as presented in Fig.~\ref{fig:cross-over}. Here, the exchange is held on the term and factor levels. We pose a number of limitations on the exchange to avoid the creation of repeating terms or the loss of derivatives inside active terms.

\begin{figure}[ht!]
\centering
\includegraphics[width=0.85\textwidth]{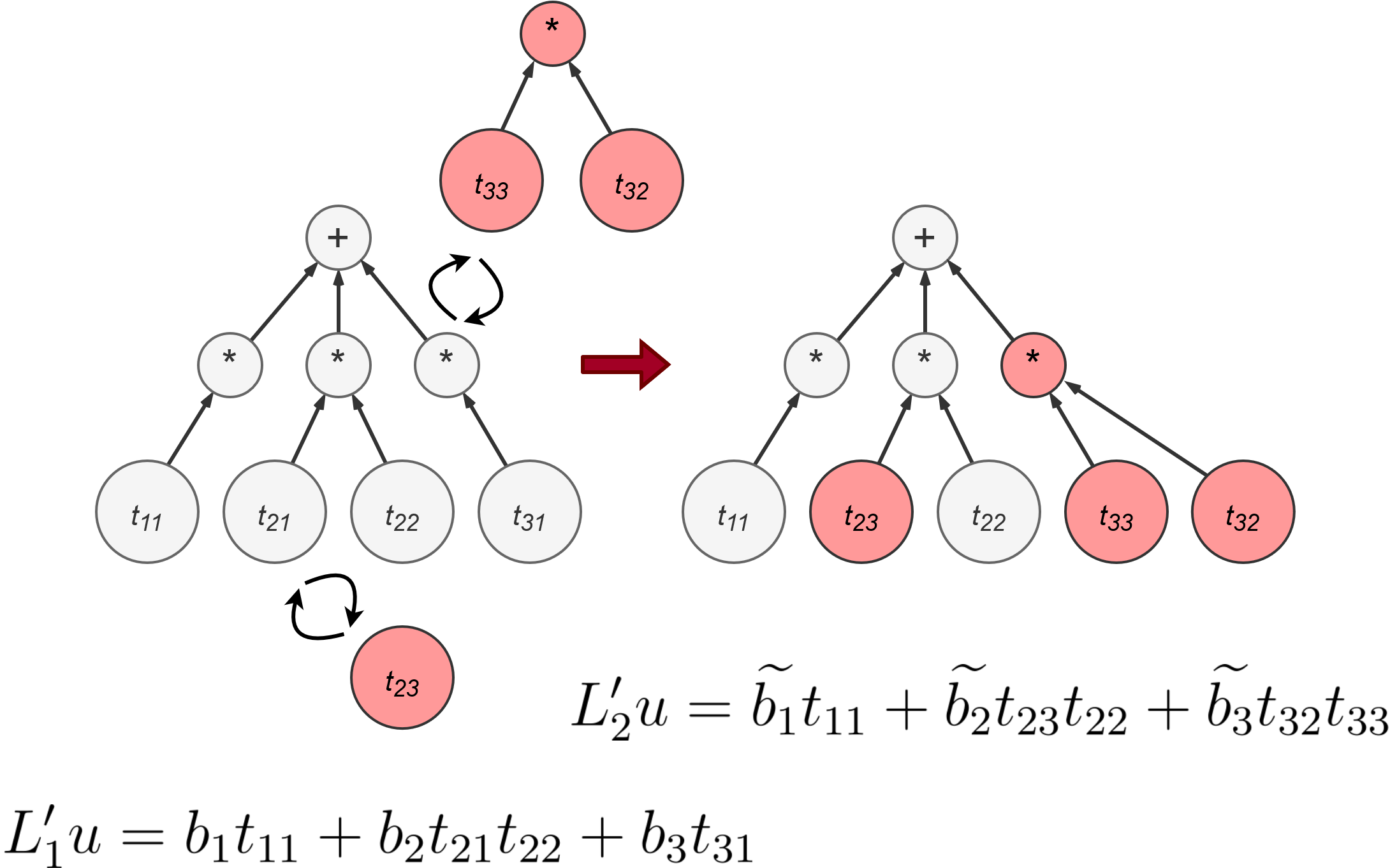}
\caption{Scheme of the mutation operator.}
\label{fig:mutation}
\end{figure}

The set of offspring created on the cross-over step is subject to mutation. The mechanics of this operation are presented in Fig.~\ref{fig:mutation}. However, if the solution obtained during the mutation is not unique in the population, it must be affected by additional mutation operations until the uniqueness is achieved. The mutation of equation construction parameters involves the addition of increments obtained from the pre-defined distribution.
 
\textit{The update procedure} involves insertion of the sub-population of new candidate solutions $P^{offspr} = \{S^{offspr}_1(\mathbf{u}), \; ... \; , S^{offspr}_{n\_offspr}(\mathbf{u}) \}$ and deletion of the less favorable ones from the population $P$ in order to preserve its size. The updating of the population in response to the offspring solution $S^{offspr}_j(\mathbf{u})$ can be held by two scenarios: if all solutions are associated with a single non-domination level, and if multiple levels exist in the population. The new solution is added to the population, and the algorithm has to delete the least feasible solution.

If there is only a single non-domination level, the algorithm tries to promote the solutions, matching the optimization sub-problem associated with the weight vectors for solutions. The algorithm detects the sub-region with the highest count of candidate solutions (or the highest sum of penalty-based intersections for solutions if there are multiple sub-problems with the same number of related solutions). It selects the system with the highest PBI value to be deleted from the population. 

In the opposite case, the algorithm examines the last non-domination level. If its cardinal number is one, then its only solution $S_{k}(\mathbf{u})$ is examined for the sub-problem region belonging: if $S_{k}(\mathbf{u})$ is the only solution, related to the subregion $\Phi_k$. It is vital for preserving population diversity and constructing the Pareto-optimal set. Thus, it is not subject to deletion and the solution with the highest PBI value is removed from the population. If there are multiple solutions associated with the subregion $\Phi_k$, then $S_{k}(\mathbf{u})$, then $S_{k}(\mathbf{u})$ can be removed. 

When multiple solutions exist in the last non-domination level, the most crowded sub-region $\Phi_k$ is subject to sparsity. The solution with the highest penalty-based intersection must be removed from the population.  

As the algorithm termination condition, the maximum number of epochs is selected: we can not pose conditions on the objective values of the systems. Equally, no conditions can be based on detecting a single non-domination level. In many cases, with a restricted number of candidate solutions, the solutions will be dispersed into a single non-domination level, and the objective of the optimization is to move the set of solutions in the direction of the ideal solution. The generalized scheme of the approach is represented in pseudocode in Algorithm~\ref{alg:MOEADD}.

\begin{algorithm}[]
\label{alg:MOEADD}
 \KwData{set of tokens $T = \{T_1, T_2, \; ...\; T_n\}$, divided into dependent and independent variables; selected optimization metrics: quality of process representation and complexity of systems; sparsity values range; values of the algorithm meta-parameters }
 \KwResult{Pareto-optimal set of differential equations/systems of differential equations} 
 Create a set of weight vectors $\mathbf{w} : w^i=(w^i_1, \; ..., \; w^i_{n\_eq + 1}) $\;
 Neighbourhood selection for sectors with respect to the acute angle \;
 Construct initial population of candidate solutions, calculate values of objective functions \& associate them with sectors \;
 Divide the initial population into non-dominated levels\;
 \For{epoch = 1 to epoch\_number}{
  \For{weight\_vector in weights}{
   Parent selection with respect to the neighbourhood of weight\_vector\;
   Apply the recombination operator to the parents' pool and apply the mutation operators to the offspring\;
   \For{offspring in new\_solutions}{
    Apply the sparsity operator and coefficient calculation,  
    calculate the criteria values for the offspring, and add it to the population \;
    }
    \uIf{the population belongs to multiple non-dominated levels}{
        Delete the solution from the last non-dominated level if the solution is not vital for population diversity else select the solution with the highest penalty-based intersection from the population from the second from last level \; 
        }
    \Else{
      Delete the solution with the highest penalty-based intersection value from the population \;
    }
   Update the population\;
  }
 }
 \caption{The pseudocode of system of equations discovery, using adapted multi-objective optimization algorithm, based on dominance and decomposition.}
\end{algorithm}

\section{Validation of the algorithm}
\label{sec:validation}

The following validation study is conducted on the varied differential equations, representing partial differential equations, an ordinary differential equation, and a system of ordinary differential equations. To better understand the strong and weak sides of the algorithm and its limitations, we have examined the performance of the SINDy (Sparse Identification of Non-linear Dynamics) framework on the same problems and compared their performance. The solutions of the equations with artificially added noise of varying magnitudes are used as the synthetic input data. Due to the stochastic nature of the evolutionary search, we have conducted ten independent runs of EPDE in each case, while SINDy operates on a more deterministic basis.

The experiments' results with sparse identifications depend on explicitly defined sparsity parameters. Therefore, a set of experiments has to be conducted with different values of the defining parameter. The models obtained can be compared to represent a validation dataset, and the best was selected as the optimal one.

The algorithm performance was analyzed from the point of view of evaluating the frequency of the algorithm and discovering the correct equations among $n$ - independent algorithm launches. The term-by-term comparison allows us to identify cases when only the most significant part of the equation was detected. As metrics, we use the empirical frequency of term detection in the structure $P, \%$, and the interval of 1.98 standard deviations from the mean value as $b, \mu \pm 1.98 \sigma$. For the SINDy framework, verifying if the equation can be correctly detected is more straightforward due to the deterministic approach. We present the resulting equation and check if it matches the expected one.

The experiments have been conducted on precise and noised datasets. In the latter case, the data was corrupted with contaminating noise $\epsilon$ as $\Tilde{u}(t, \mathbf{x}) = u(t, \mathbf{x}) + \epsilon(t, \mathbf{x})$ obtained from the normal distribution $\epsilon(t, \mathbf{x}) \sim \mathcal{N}(0, k \cdot u(t, \mathbf{x}))$, and introduced into the dataset. To reduce the uncertainty of comparing experiments with different frameworks, we estimate it with the noise level metric $NL, \%$ introduced as in Eq.\eqref{eq:noise_level}.

\begin{equation}
NL = \frac{\vert  u(t, \mathbf{x}) - \Tilde{u}(t, \mathbf{x}) \vert }{\vert  u(t, \mathbf{x}) \vert}
\label{eq:noise_level}
\end{equation}

The contaminating distribution parameters were selected accordingly to obtain the desired noise level. The table in \ref{app:params} summarizes the framework setup for each experiment for each experiment.

\subsection{Partial differential equations}
\label{subsec:PDE_validation}

\subsubsection{Burgers' equation}

First, we examine the case of partial differential equation discovery. Methods based on sparse regression among second-order PDEs can operate only with parabolic-like equations. The first case to be examined is Burgers' equation Eq.\eqref{eq:Burgers}. It represents the momentum balance for the one-dimensional Navier-Stokes equation, with $\nu$ representing the viscosity. With the viscosity values close to zero, the equation devolves into the non-linear homogeneous first-order PDE $u'_{t} + u \cdot u'_{x} = 0$.

\begin{equation}
\label{eq:Burgers}
    \frac{\partial u}{\partial t} + u \frac{\partial u}{\partial x} = \nu \frac{\partial^2 u}{\partial x^2}
\end{equation}

In this example, we will study the system with parameter $\nu = 0.1$, provided by SINDy developers. The dataset comprises 101 time points and 256 spatial points with steps of 0.1 and 0.0625, respectively.

\begin{table}[h!]
\caption{Correct terms inclusion statistics for the Burgers' equation and of the corresponding coefficients, paired with obtained by SINDy for the specified noise levels. The abbreviation g.t. denotes ground truth.}
\tiny
\begin{center}
 \begin{tabular}{| c | c | c | c | c | c | c | c |} 
  \hline
  \multirow{3}{*}{NL, $\%$} & \multicolumn{6}{|c|}{EPDE} & \multirow{2}{*}{SINDy}   \\ \cline{2-7}
  & \multicolumn{2}{|c|}{$u'_{t}$} & \multicolumn{2}{|c|}{$u''_{xx}$} &  \multicolumn{2}{|c|}{$u u'_{x}$} &    \\ \cline{2-8}
  & $P, \%$ & $b, \mu \pm 1.98 \sigma$ & $P, \%$ & $b, \mu \pm 1.98 \sigma$ & $P, \%$ & $b, \mu \pm 1.98 \sigma$ & g.t. $u'_t=0.1u''_{xx}-uu'_x$  \\ 
  \hline
  0 & $100$ & $1.001 \pm 0$  & $100$ & $0.106 \pm 0.0$ & $100$ & $-0.997 \pm 0.0$ & $u'_{t} = 0.1 u''_{xx} - 1.001 u u'_{x}$   \\ 
  \hline  
  1 & $90$ & $0.830 \pm 0.218$  & $60$ & $0.053 \pm 0.002$ & $10$ & $-0.980 \pm 0.0$ & $u'_{t} = -0.248 u'_{x} - 0.292 u u'_{x}$   \\
  \hline
  2.5 & $80$ & $0.599 \pm 0.158$ & $50$ & $0.018 \pm 0.0$ & $0$ & $-$ & $  u'_{t}=-0.265 u'_{x} - 0.229 u u'_{x}$   \\
  \hline
  5 & $100$ & $0.674 \pm 0.139$ & $20$ & $-0.012 \pm 0.0$ & $0$ & $-$  & $u'_{t}=- 0.001 u u'''_{xxx}-0.825 u u'_{x}  $   \\  
  \hline
  10 & $100$ & $0.674 \pm 0.103$  & $40$ & $0.004 \pm 0.0$ & $0$ & $-$ & $ u'_{t}=0.133 u u''_{xx}$  \\  
  \hline
\end{tabular}
\label{tab:burgers_res}
\end{center}
\end{table}

The execution time for the EPDE framework runs 91 seconds on average, while a search with SINDy takes 0.032 seconds. This time discrepancy can be attributed to the algorithmic simplicity of the execution and the smaller search space. The validation results are summarized in Tab.~\ref{tab:burgers_res}. With the introduction of noise, both approaches rapidly lose the ability to derive the equation with the correct structures. The algorithm can reliably converge to the correct equation with the correct coefficients only with noise levels equal to or lower than $1\%$.

\subsubsection{Korteweg-de Vries equation}

The Korteweg-de Vries equation is an example of a more complex partial differential equation that includes tokens of high-order derivatives in its structure. It is a non-linear partial differential equation of the third order, and here, we will examine its homogeneous case. The equation is presented in Eq.~\eqref{eq:Korteweg-de_Vries} and represents the transition of a solitary wave with constant velocity.

\begin{equation}
\label{eq:Korteweg-de_Vries}
    \frac{\partial u}{\partial t} + 6 u \frac{\partial u}{\partial x} + \frac{\partial^3 u}{\partial x^3} = 0
\end{equation}

We have conducted experiments on the data set, which contains 512 temporary and 201 spatial points, matching the time steps of $0.1$ and $\approx 0.12$. In this case, the data represent the traversal of two solitary waves.

\begin{table}[h!]
\caption{Correct terms inclusion statistics for the Korteweg-de Vries equation and of the corresponding coefficients, paired with obtained by SINDy for the specified noise levels. Abbreviation g.t. denotes ground truth and $N[u]= -0.515 u'_{x}+3.813 u^2 u'_{x} - 0.013 u u'''_{xxx} + 0.025 u^2  u'''_{xxx}$.}
\tiny
\begin{center}
 \begin{tabular}{| c | c | c | c | c | c | c | c |} 
  \hline
  \multirow{3}{*}{NL, $\%$} & \multicolumn{6}{|c|}{EPDE} & \multirow{2}{*}{SINDy}   \\ \cline{2-7}
  & \multicolumn{2}{|c|}{$u'_{t}$} & \multicolumn{2}{|c|}{$u u'_{x}$} & \multicolumn{2}{|c|}{$u'''_{xxx}$}  &    \\ \cline{2-8}
  & $P, \%$ & $b, \mu \pm 1.98 \sigma$ & $P$ & $b, \mu \pm 1.98 \sigma$ & $P$ & $b, \mu \pm 1.98 \sigma$ & g.t. $u'_t+u'''_{xxx}+6uu'_x=0$    \\ 
  \hline
  0 & $100$ & $1.001 \pm 0.0$ & $100$ & $6.002 \pm 0.0$ & $100$ & $1.06 \pm 0.0$ & $u'_{t} + 0.992 u'''_{xxx} + 5.967 u u'_{x} = 0$   \\ 
  \hline
  0.5 & $80$ & $0.913 \pm 0.032$ & $60$ & $5.914 \pm 2.59$ & $70$ & $1.31 \pm 0.57$ & $u'_{t}- 0.906 u'_{x} = 0$   \\
  \hline
  1 & $40$ & $0.437 \pm 0.156$ & $0$ & $-$ & $0$ & $-$ & $u'_{t}-0.816 u'_{x} = 0$   \\
  \hline
  2.5 & $100$ & $0.36 \pm 0.0$ & $20$ & $1.0 \pm 0.0$ & $20$ & $0.01 \pm 0.0$ & $u'_{t} - 0.004 u'''_{xxx}-0.844 u'_{x} = 0$   \\  
  \hline
  5 & $60$ & $0.01 \pm 2.13 \cdot 10^{-5}$ & $80$ & $1.0 \pm 0.0$ & $0$ & $-$ & $ u'_{t} - 0.003 u'''_{xxx} -1.859 u u'_{x}+N[u]=0$  \\  
  \hline
\end{tabular}
\label{tab:kdv_res}
\end{center}
\end{table}

The results of the experiments are presented in Tab.~\ref{tab:kdv_res}. Despite the correct terms found by SINDy for the 5\% case, it is hardly recovered from the provided discovered equation due to an excessive number of terms $N[u]$ with higher coefficients. Due to the presence of the third-order derivative, the correct structure can be lost on the relatively low noise level if the finite-difference method is used. Using data smoothing and fitting of Chebyshev polynomials and evolutionary search, the differentiation method achieves convergence even on noise levels greater than $0.5\%$ in some runs. The search time with the evolutionary approach took 72 seconds for SINDy - 0.041 seconds.

\subsection{Ordinary differential equations}
\label{subsec:ODE_validation}

The ability of the proposed framework to derive ordinary differential equations can be demonstrated on the Van der Pol oscillator. Initially introduced to describe the relaxation-oscillation cycle produced by the electromagnetic field, the model has found applications in other spheres of science, such as biology or seismology. The model takes the form of an ordinary differential equation of the second order as in Eq.~\eqref{eq:van_der_pol_eq} with $\mathcal{E}$ - positive constant (in the experiments, $\mathcal{E} = 0.2$).

\begin{equation}
u'' + \mathcal{E}(u^2 - 1)u' + u = 0
\label{eq:van_der_pol_eq}
\end{equation}

Although the proposed evolutionary approach can handle the approach directly, the equation can be transformed into a system of two first-order differential equations Eq.~\eqref{eq:van_der_pol_sys_eq}, which in theory can be processed with a sparse identification of dynamics.

\begin{equation}
\begin{cases}
    u' = v; \\
    v' = -\mathcal{E}(u^2 - 1)v - u
\end{cases}
\label{eq:van_der_pol_sys_eq}
\end{equation}

The dataset included the solution of the equation with initial conditions of $u = \sqrt{3}/2; u' = 1/2$ for a domain of 320 points with the step of $0.05$ starting from a conventional point $t=0$. The numerical solution was computed using the Runge-Kutta method of order 4.

The challenge of applying an evolutionary-based approach to the problem in the form of an ordinary differential equation decomposed into a system of differential equations arises in the presence of trivial forms related to the equality of tokens $x' = y$, guided by the first equation of the system, and to the versatility of the algorithm. During the evolution, the algorithm proposes candidate equations, such as $v \cdot u' \cdot v' = v' (u')^2$, which are trivial identities concerning the correctly discovered equation $u' = v$.

Despite the inability to obtain an ODE as a system of first-order equations, the proposed algorithm correctly identifies the correct model in the single-equation mode. Analysis of the predictions, based on the obtained equations, can give insight into how errors in equation detection affect the predicting ability of the model. 

\begin{figure}[ht!]
\centering
\includegraphics[width=0.90\textwidth]{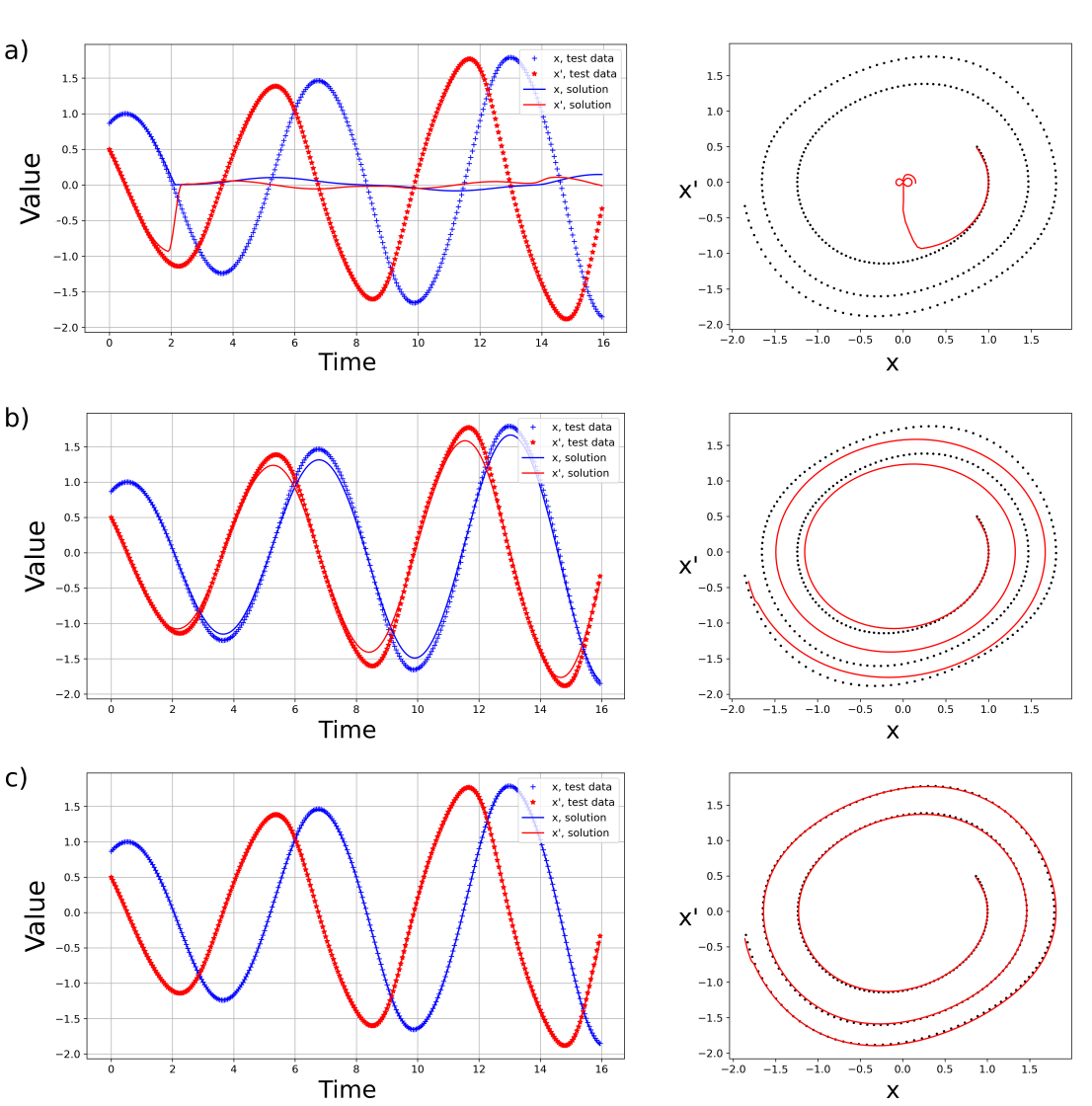}
\caption{Example of Van der Pol oscillator state predictions, based on obtained differential equations. a) Equation with misidentified structure; b) Equation with correct terms but deviating coefficients; c) Predictions with the correct data-driven equation.} 
\label{fig:VdP_pred}
\end{figure}

An illustration of the algorithm performance test on the time interval adjacent to the training one is presented in Fig.~\ref{fig:VdP_pred}. Case (a) represents an example of a prediction with an entirely incorrect equation structure, amplified by an undertrained predicting neural network. In particular, this case shows that the validation data set must be sufficiently long to represent the process-representing properties of the equations. With both the evaluation of the fitness of the equation during optimization and the validation involving the solution of the initial value problem, the error is significantly lower in the domain next to the defined conditions. Here, in section $[0, \approx 1.7]$, the solution of the proposed equation does not significantly deviate from the correct values.

The following case (b) represents the predictions with data-driven differential equations with the correct structure, while the coefficients are evaluated with marginal error. In case the properties of a discovered dynamical system, linked to the equation, with a vector of coefficients $\mathbf{\alpha}'$ do not lead to bifurcation from the solution of the expected system with parameters $\mathbf{\alpha}$, the candidate tends to provide feasible predictions. The solution of data-driven equations with minimal deviations from the actual coefficients is presented in part (c) of Fig.~\ref{fig:VdP_pred} and follows the expected dynamics with minimal inaccuracies. 

\begin{table}[h!]
\caption{Correct terms inclusion statistics for the equation, describing Van der Pol oscillator dynamics. Here by ground truth we denote the equation form of $u'' = - 0.2(u^2 - 1)u' - u$.}
\tiny
\begin{center}
 \begin{tabular}{| c | c | c | c | c | c | c | c | c |} 
  \hline
  \multirow{3}{*}{NL, $\%$} & \multicolumn{8}{|c|}{EPDE} \\ \cline{2-9}
  & \multicolumn{2}{|c|}{$u^2 u'_{t}$} & \multicolumn{2}{|c|}{$u'_{t}$} & \multicolumn{2}{|c|}{$u$} & \multicolumn{2}{|c|}{$u''_{tt}$} \\ \cline{2-9}
  & $P, \%$ & $b, \mu \pm 1.98 \sigma$ & $P, \%$ & $b, \mu \pm 1.98 \sigma$ & $P$ & $b, \mu \pm 1.98 \sigma$ & $P$ & $b, \mu \pm 1.98 \sigma$ \\ 
  \hline
  0 & $100$ & $-0.199 \pm 0.0$ & $100$ & $0.199 \pm 0.0$ & $100$ & $1.00 \pm 0.0$ & $100$ & $1.00 \pm 0.0$    \\ 
  \hline
  0.5 & $100$ & $-0.181 \pm 33.6 \cdot 10^{5}$ & $100$ & $0.187 \pm 11.9 \cdot 10^{5}$ & $100$ & $1.00 \pm 5.9 \cdot 10^{5}$ & $100$ & $1.0 \pm 0.0$  \\
  \hline
  1 & $60$ & $0.085 \pm 3.96 \cdot 10^{5}$ & $0$ & $-$ & $80$ & $0.065 \pm 1.98 \cdot 10^{5}$ & $0$ & $-$   \\
  \hline
  2.5 & $100$ & $0.36 \pm 0.0$ & $20$ & $1.0 \pm 0.0$ & $20$ & $0.01 \pm 0.0$ & $0$ & $-$  \\  
  \hline
  5 & $60$ & $0.013 \pm 1.63 \cdot 10^{-5}$ & $80$ & $1.0 \pm 0.0$ & $0$ & $-$ & $0$ & $-$  \\  
  \hline
\end{tabular}
\label{tab:vdp_results_epde}
\end{center}
\end{table}

\begin{table}[h!]
\caption{Equations, obtained by applying SINDy algorithm to identify the equation, describing Van der Pol oscillator. As the ground truth (g.t.) in this case, we define the system of first-order ODEs, equivalent to the initial equation.}
\small
\begin{center}
 \begin{tabular}{| c | c |} 
  \hline
  NL, $\%$ & SINDy equations  \\
  \hline
  g.t. & $\begin{cases} u' = v ; \\ v' = 0.2 \cdot (u^2 - 1)v - u \end{cases}$   \\ 
  \hline  
  0 & $\begin{cases} u' = 1.004v - 0.001 u^2 v; \\ v' = -0.962u - 0.017u^3 - 0.016 u v^2 \end{cases}$   \\
  \hline
  0.5 & $\begin{cases} u' = 0.996v + 0.007 v u^2 - 0.003 v^3; \\ v' = -0.881 u + 0.551 u^2 + 0.566 u v - 0.003 u^4 - 0.482 u^3 v \end{cases}$   \\
  \hline
  1 & $\begin{cases} u' =  1.004v - 0.001 u^2 v + 0.001 v^3; \\ v' = -1.243 u + 0.342 u v + 0.137 u^{3} + 0.059 u v^2 - 0.297 u^3 v \end{cases}$   \\
  \hline
  2.5 & $\begin{cases} u' =  1.05 v - 0.031 v^3; \\ v' = -0.72 u + 0.612 u v + 0.043 u^{3} + 0.163 u^2 v - 0.286 u v^2  + \\ + 0.095 u^3 v - 0.127 u^2 v^2 - 0.465 u v^3 \end{cases}$   \\  
  \hline
  5 & System was not discovered due to convergence issues  \\
  \hline
\end{tabular}
\label{tab:vdp_results_sindy}
\end{center}
\end{table}

The Van der Pol oscillator case presents an interesting test case due to the high complexity of the governing equations. Statistics presented in Tab.~\ref{tab:vdp_results_epde} for the proposed evolutionary approach, and in Tab.~\ref{tab:vdp_results_sindy} show that up to the noise levels of $1\%$, the evolutionary-based approach can correctly derive the equation. Then, even in cases of the presence of three of four terms, the coefficients are incorrectly identified. Sparse identification can detect the introduction of a variable $v = u'$ with insignificant noise-related parts. However, it cannot correctly form the equation for $v'$. The average equation search with sparse identification took 0.015 seconds of runtime, while with the evolutionary approach the execution took an average of 48 seconds (without the procedure of derivative calculation; with it - 458 seconds, although the number of ANN training iterations has proved excessive).

\subsection{Systems of differential equations}
\label{subsec:system_validation}

Next, we shall discuss the performance of the proposed algorithm on problems of discovering differential equation systems. A simple example of such a system is the hunter-prey model, also known as the Lotka-Volterra system, presented in Eq.~\eqref{eq:lotka_volterra}. It describes the dynamics of a primitive ecosystem with a single hunter species with population size $v$ and a single prey of size $u$. Non-negative parameters $\alpha$, $\beta$, $\delta$, and $\gamma$ control the behavior of the system.  

\begin{equation}
\label{eq:lotka_volterra}
\begin{cases}
    \frac{d u}{d t} = \alpha u - \beta u v; \\
    \frac{d v}{d t} = \delta u v - \gamma v;
\end{cases}
\end{equation}

We used the system with the following parameters in the experiment: $\alpha = \beta = \gamma = \delta = 20$. The initial condition required by the system was set as $u = 4, \; v = 2$, and the equations were numerically solved with the Runge-Kutta fourth-order method.

The evaluated non-linear system has a relatively simple yet illustrative structure and can be discovered by an evolutionary approach and sparse identification. 

\begin{table}[h!]
\caption{Correct terms inclusion statistics for the equation, describing prey dynamics, with the corresponding coefficients, and the equation obtained by SINDy. Ground truth equation is denoted by abbreviation g.t., and $N_{NL}[u, v]$ - additional less significant terms, detected by SINDy in equation structure.}
\tiny
\begin{center}
 \begin{tabular}{| c | c | c | c | c | c | c | c |} 
  \hline
  \multirow{3}{*}{NL, $\%$} & \multicolumn{6}{|c|}{EPDE} & \multirow{2}{*}{SINDy}   \\ \cline{2-7}
  & \multicolumn{2}{|c|}{$u$} & \multicolumn{2}{|c|}{$u'$} & \multicolumn{2}{|c|}{$uv$}  &    \\ \cline{2-8}
  & $P, \%$ & $b, \mu \pm 1.98 \sigma$ & $P$ & $b, \mu \pm 1.98 \sigma$ & $P$ & $b, \mu \pm 1.98 \sigma$ & g.t. $u' = 20 u - 20 u v$  \\ 
  \hline
  0 & $100$ & $19.83 \pm 0.24$ & $100$ & $1.0 \pm 0.0$ & $90$ & $-20.06 \pm 0.008$ & $u' = 20.096 u - 19.842 u v + N_{0}[u, v]$   \\ 
  \hline
  0.5 & $100$ & $19.969 \pm 0.0$ & $100$ & $1.0 \pm 0.0$ & $100$ & $-20.214 \pm 0.0$ & $u' = 20.194 u - 19.87 u v + N_{0.5}[u, v]$   \\
  \hline
  1 & $90$ & $19.070 \pm 0.263$ & $100$ & $1.0 \pm 0.0$ & $40$ & $-19.361 \pm 0.0$ & $u' = 20.726 u - 19.904 u v + N_{1.0}[u, v]$   \\
  \hline
  2.5 & $50$ & $6.964  \pm 175.4$ & $60$ & $0.38 \pm 0.368$ & $10$ & $1.4 \pm 0.0$ & $u' = 19.311 u - 19.67 u v + N_{2.5}[u, v]$   \\  
  \hline
  5 & $30$ & $-2.77 \pm 39.3$ & $50$ & $0.1 \pm 0.011$ & $10$ & $1.4 \pm 0.0$ & Convergence failure  \\  
  \hline
\end{tabular}
\label{tab:lotka_u}
\end{center}
\end{table}

\begin{table}[h!]
\caption{Correct terms inclusion statistics for the equation, describing hunter dynamics, with the corresponding coefficients and equations, obtained by SINDy. Ground truth equation is denoted by abbreviation g.t., and $N_{NL}[u, v]$ - additional less significant terms, detected by SINDy in equation structure.}
\tiny
\begin{center}
 \begin{tabular}{| c | c | c | c | c | c | c | c |} 
  \hline
  \multirow{3}{*}{NL, $\%$} & \multicolumn{6}{|c|}{EPDE} & \multirow{2}{*}{SINDy}   \\ \cline{2-7}
  & \multicolumn{2}{|c|}{$v$} & \multicolumn{2}{|c|}{$u'$} & \multicolumn{2}{|c|}{$uv$}  &    \\ \cline{2-8}
  & $P, \%$ & $b, \mu \pm 1.98 \sigma$ & $P$ & $b, \mu \pm 1.98 \sigma$ & $P$ & $b, \mu \pm 1.98 \sigma$ & g.t. $v' = - 20 v + 20 u v$  \\ 
  \hline
  0 & $90$ & $-20.018 \pm 0.0$ & $90$ & $1.0 \pm 0.0$ & $90$ & $24.741 \pm 384.9$ & $v' = -19.97 v + 19.85 u v + N_{0.5}[u, v]$   \\ 
  \hline
  0.5 & $100$ & $-19.822 \pm 0.0$ & $100$ & $1.0 \pm 0.0$ & $100$ & $20.098 \pm 0.0$ & $v' = -20.99 v + 19.86 u v + N_{0.5}[u, v]$  \\
  \hline
  1 & $100$ & $-19.922 \pm 33.6 \cdot 10^{-4}$ & $100$ & $1.0 \pm 0.0$ & $100$ & $20.011 \pm 0.021$ & $v' = - 19.73 v + 19.63 u v - N_{1.0}[u, v]$   \\
  \hline
  2.5 & $90$ & $-18.987 \pm 1.09$ & $90$ & $1.0 \pm 0.0$ & $40$ & $31.26 \pm 816.2$ & $v' = - 20.65 v - 20.12 u v + N_{2.5}[u, v]$   \\  
  \hline
  5 & $40$ & $-8.97 \pm 65.0$ & $50$ & $0.525 \pm 0.28$ & $70$ & $72.86 \pm 25.7$ & Convergence failure  \\  
  \hline
\end{tabular}
\label{tab:lotka_v}
\end{center}
\end{table}

The results of the equation discovery are illustrated in Tab.~\ref{tab:lotka_u} and Tab.~\ref{tab:lotka_v}. By $\epsilon$, we denote the insignificant terms in the equations derived from the SINDy framework. From the noise levels of $1\%$ and higher, the sparse identification has not converged to a single equation. The evolutionary-based approach can reliably distil the correct governing equation on noise levels up to $5\%$, while no attempts have resulted in success on higher noise levels. The equation search took 68 seconds on average with the evolutionary approach and 0.032 seconds with the SINDy framework.

\section{Experimental summary}
\label{sec:exp}

The experimental results are summarized in Tab.~\ref{tab:exp_concl}. The central aggregated values are the frequency of successfully obtaining the correct equations in the experiments and the errors of the calculated coefficients over all experiments shown above. In the summarizing table, in columns of positive discoveries (Pos., $\%$), we present the frequencies of the correct equation on data with the noise of value in the noise magnitude column (noise level, NL $\%$). For the evolutionary approach, even a single successful equation discovery can be beneficial and prove that the algorithm can converge to the correct structure. Although the correct structure may not be optimal on noisy training data, it can be identified on the held-out sample so that additional data can be used for model rectification. Thus, the frequency column of one positive equation discovery is introduced in the results (one pos. in exp., $\%$).  

\begin{table}[h!]
\caption{Experiments statistics: frequency-based comparison of obtaining the correct equations and the coefficient calculation errors.}
\tiny
\begin{center}
 \begin{tabular}{| c | c | c | c | c | c |} 
  \hline 
   \multirow{2}{*}{NL $\%$} & \multicolumn{2}{|c|}{SINDy} & \multicolumn{3}{|c|}{EPDE}    \\ \cline{2-6} %
   & Pos., $\%$ & Coeff. error MAPE $\%$ & Pos., $\%$ & One pos. in exp., $\%$ & Coeff. error MAPE $\%$, best \\ 
  \hline
  0 & \cellcolor{green!10}100 & \cellcolor{red!10}0.3 & \cellcolor{red!10}97.5 & \cellcolor{green!10}100 & \cellcolor{green!10}0.21 \\ 
  \hline
  0.5 & \cellcolor{red!10}25 & \cellcolor{green!10}0.95 & \cellcolor{green!10}67.5 & \cellcolor{green!10}100 & \cellcolor{red!10}7.5 \\ 
  \hline
  1 & \cellcolor{yellow!10}25 & \cellcolor{red!10}1.46 & \cellcolor{red!10}10 & \cellcolor{yellow!10}25 & \cellcolor{green!10}1.19 \\ 
  \hline
  2.5 & \cellcolor{yellow!10}25 & \cellcolor{red!10}1.49 & \cellcolor{red!10}2.5 & \cellcolor{yellow!10}25 & \cellcolor{green!10}1.31 \\ 
  \hline
  5 & \cellcolor{red!10}0 & \cellcolor{red!10}- & \cellcolor{green!10}2.5 & \cellcolor{green!10}25 & \cellcolor{green!10}4.9 \\ 
  \hline
\end{tabular}
\label{tab:exp_concl}
\end{center}
\end{table}

Reliable detection with a sparse approach can be expected only on noiseless data. Only the Lotka-Volterra system can be detected at higher noise levels due to its simple structure and the absence of high-order derivatives. An evolutionary approach can be applied to noisy data, even though high noise contamination leads to the wrong equation being optimal from an optimization criteria standpoint.

In the column for coefficient error assessment, we compare the ones that match the correct terms of the correctly identified equations. Here, the superior data preprocessing procedures in EPDE allow for more precise coefficient calculation. 

\section{Conclusion}
\label{sec:concl}

In this study, we have explored the opportunities to employ an evolutionary approach to data-driven differential equation discovery instead of conventional sparsity promotion. The main results of the comparison can be summarized in the following points:

\begin{itemize}
    \item The evolutionary method achieves its primary objective of providing a solution to the central issue of sparse regression: to adapt the term library to the problem. Using elementary functions as the building blocks of the equations allows the factors of the equation to contain parameters subject to optimization. This advantage can be vital in the case of unknown dependent variable representations. However, the sparse identification method can be optimal in some cases due to the issues linked with the increased computational cost of getting the model and the consequent longer optimization time. Such scenarios include those we can be sure of in the presence of time dynamics in the form of first-order derivatives.  

    \item Although for the integrity of the comparison, we have examined only the first-order partial differential equations from the time axis point that can be discovered by sparse identification, it is necessary to point out the abundance of equations without the first partial derivative along the time axis. Fundamental cases such as Poisson and the wave equation do not contain $u'_{t}$ and, therefore, cannot be discovered with sparse regression. With ordinary differential equations, when we expect higher-order derivatives, we can introduce additional variables representing the derivatives of the main variable. 
    
    \item In the evolutionary approach, significant attention shall be paid to selecting the preprocessing tool, i.e. noise-resistant derivative calculation method, and the fitness function.
    Correctly selecting these elements can significantly increase the noise threshold to derive the correct equations, albeit further increasing the demand for the required calculations.   
    
    \item The automated control of the equation discovery parameters can be done using a multi-objective evolutionary optimization approach. With its help, the researcher can explore the complexity-quality trade-off among the candidate differential equations on the Pareto-optimal set.
\end{itemize}

Despite all advances, some improvements must be made to increase the applicability of data-driven differential equation discovery methods to real-world problems. First, the developed approach requires manually setting multiple parameters and case-specific elementary functions. This issue can discourage potential users or lead to non-optimal equation discovery if these settings are chosen incorrectly. Another challenge is linked to high execution time compared to conventional machine learning techniques. Developing computationally cheaper alternatives to the provided preprocessing operators or fitness functions has to be one of the priority research areas. 

\section*{Code and Data availability}
\label{sec:repo}
The Python code to reproduce the experiments and the necessary datasets are available in the GitHub repository \url{https://github.com/ITMO-NSS-team/EPDE_paper_experiments}.

\section*{Acknowledgements}

This work was supported by the Analytical Center for the Government of the Russian Federation (IGK 000000D730321P5Q0002), agreement No. 70-2021-00141.

\bibliographystyle{elsarticle-num} 
\bibliography{references}
\appendix

\pagebreak

\section{Experiment parameters}
\label{app:params}

\begin{table}[h!]
\caption{Parameters values, and module approaches, used during the experiments. The setups were not altered during the runs on noised data to present realistic scenarios of equation discovery, where noise parameters cannot be estimated.}
\begin{center}
 \begin{tabular}{| c | c | c | c | c | c |} 
 \hline
  & & Burgers eq. & KdV eq. & V.d.P. oscill. & L.-V. system  \\ [0.5ex]
  \hline 
  \multirow{7}{*}{EPDE} & Token fam. & \begin{tabular}{@{}c@{}c@{}c@{}}Deriv.\\ord=(2, 3),\\ Coord., \\ Sine$\setminus$Cosine \end{tabular} & \begin{tabular}{@{}c@{}c@{}c@{}}Deriv.\\ord=(2, 3),\\ Coord., \\ Sine$\setminus$Cosine \end{tabular} & \begin{tabular}{@{}c@{}c@{}c@{}}Deriv.\\ ord=2, \\ Coordinate \end{tabular} & \begin{tabular}{@{}c@{}c@{}c@{}}Deriv.\\ ord=2, \\ Coordinate \end{tabular} \\ \cline{2-6}
  & Preproc. & Chebyshev & Chebyshev & ANN repr. & ANN repr.  \\ \cline{2-6}
  & Qual. crit. & sol.-based & eq. discr. & eq. discr. & eq. discr.  \\ \cline{2-6}  
  & Pop. size & 8 & 8 & 12 & 12  \\ \cline{2-6}
  & Evo. epochs & 65 & 55 & 85 & 85 \\ \cline{2-6}
  & Factors num. & 2 & 2 & 4 & 2 \\ \cline{2-6}
  & Eq. len. & 5 & 6 & 6 & 5 \\ \cline{2-6}
  & Sparsity int. & $(10^{-9}, 10^{0})$ & $(10^{-9}, 10^{0})$ & $(10^{-12}, 10^{-4})$ & $(10^{-12}, 10^{-4})$ \\ \cline{2-6}
  
 \hline
 \multirow{2}{*}{SINDy} & Library & \begin{tabular}{@{}c@{}c@{}c@{}c@{}} Polynom.\\ ord.=2, \\ PDELibrary \\ ord. = 3 \end{tabular} & \begin{tabular}{@{}c@{}c@{}c@{}c@{}} Polynom.\\ ord.=4, \\ PDELibrary \\ ord. = 3 \end{tabular} & \begin{tabular}{@{}c@{}c@{}c@{}c@{}} Polynom.\\ ord.=4, \\ deriv. \\ ord. = 3 \end{tabular} & \begin{tabular}{@{}c@{}c@{}c@{}c@{}} Polynom.\\ ord.=2, \\ deriv. \\ ord. = 3 \end{tabular} \\ \cline{2-6}
  & Optimizer & SSR & SR3 & STLSQ & STLSQ  \\ \cline{2-6}
 \hline
\end{tabular}
\label{tab:param_table}
\end{center}
\end{table}




\end{document}